\documentclass[runningheads]{llncs}

\usepackage{eccv}
\usepackage{eccvabbrv}

\usepackage{graphicx}
\usepackage{booktabs}
\usepackage{multirow}
\usepackage{adjustbox} 
\usepackage{subcaption}
\usepackage{float}
\usepackage{makecell}
\usepackage[accsupp]{axessibility}  

\usepackage{hyperref}

\usepackage{orcidlink}

\begin{document}

\title{Seeing What Matters: Perceptual Wrapper with Common Randomness for 3D Gaussian Splatting}

\titlerunning{Seeing What Matters}

\author{He-Bi~Yang\inst{1} \and
Jing-Zhong~Chen\inst{1} \and
Yen-Kuan~Ho\inst{1} \and
Sang~NguyenQuang\inst{1} \and
Fan-Yi~Hsu\inst{1} \and
Yun-Yu~Lee\inst{1} \and
Jui-Chiu~Chiang\inst{2} \and
Wen-Hsiao~Peng\inst{1}
}

\authorrunning{H.-B.~Yang et al.}

\institute{National Yang Ming Chiao Tung University, Taiwan \and
National Chung Cheng University, Taiwan
}

\maketitle

\begin{abstract}
  While 3D Gaussian Splatting (3DGS) achieves impressive real-time rendering, it frequently struggles to synthesize high-frequency textures, a limitation heavily exacerbated in memory-constrained and rate-distortion-optimized (RDO) pipelines. To address this, we propose a versatile 2D perceptual wrapper that enhances the rendered outputs of existing 3DGS representations in a content- and view-dependent manner. Our method leverages a lightweight synthesis network conditioned on pseudo-random Gaussian noise to synthesize perceptually plausible textures. Supervised by Wasserstein Distortion, the network learns to match local feature statistics rather than strictly enforcing pixel-wise reconstruction fidelity, effectively mitigating the blurriness inherent in standard frameworks. We demonstrate the broad applicability of our plug-and-play approach across vanilla, memory-constrained, and RDO 3DGS methods. Comprehensive subjective and objective experiments confirm that our method significantly improves over existing baselines, yielding superior perceptual quality at sharply reduced file or model sizes.
\end{abstract}

\section{Introduction}
\label{sec:intro}
\begin{figure}[p]
    \centering
    \begin{subfigure}{\linewidth}
        \centering
        \includegraphics[width=\linewidth]{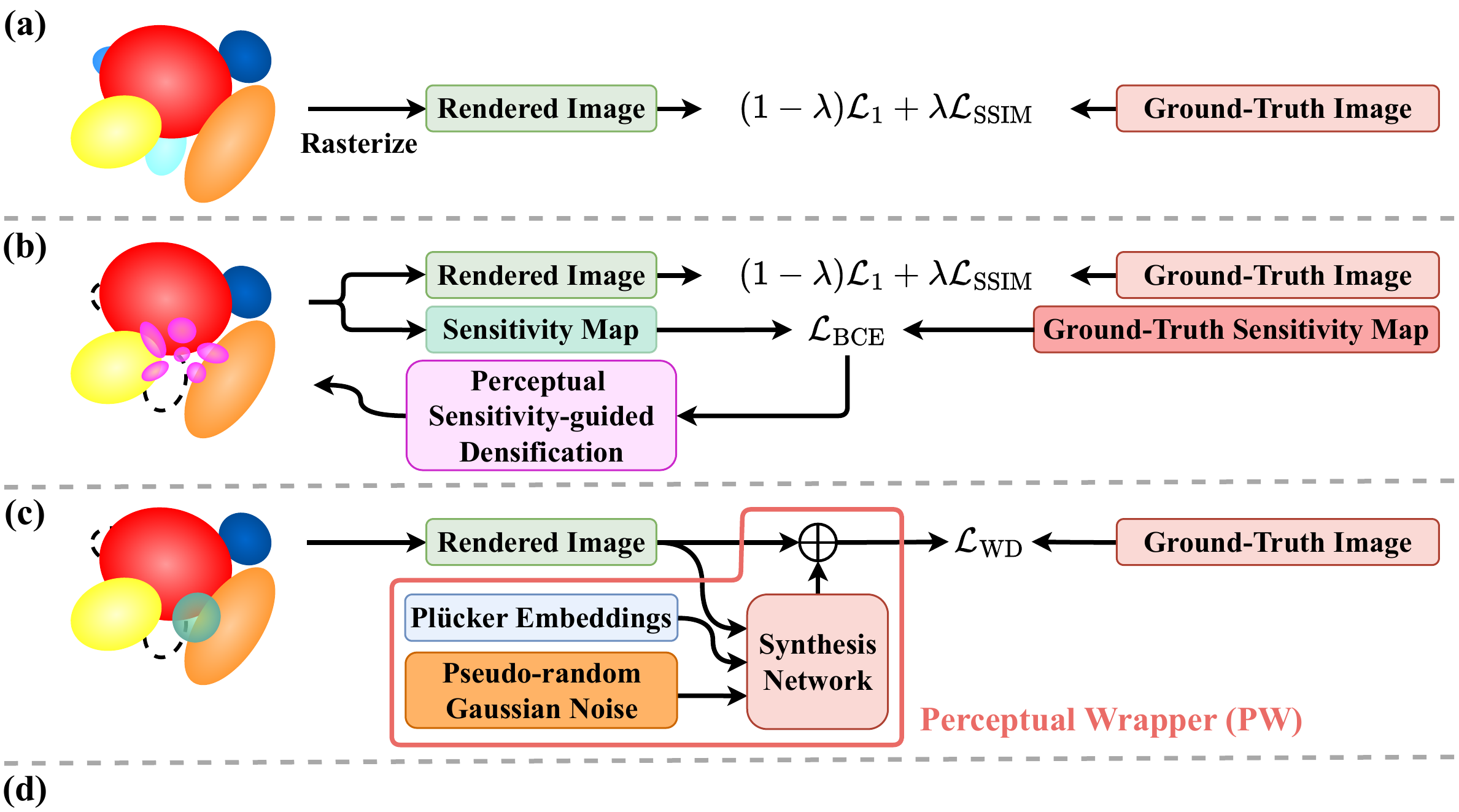}
    \end{subfigure}

    \begin{subfigure}{\linewidth}
        \centering
        \includegraphics[width=\linewidth]{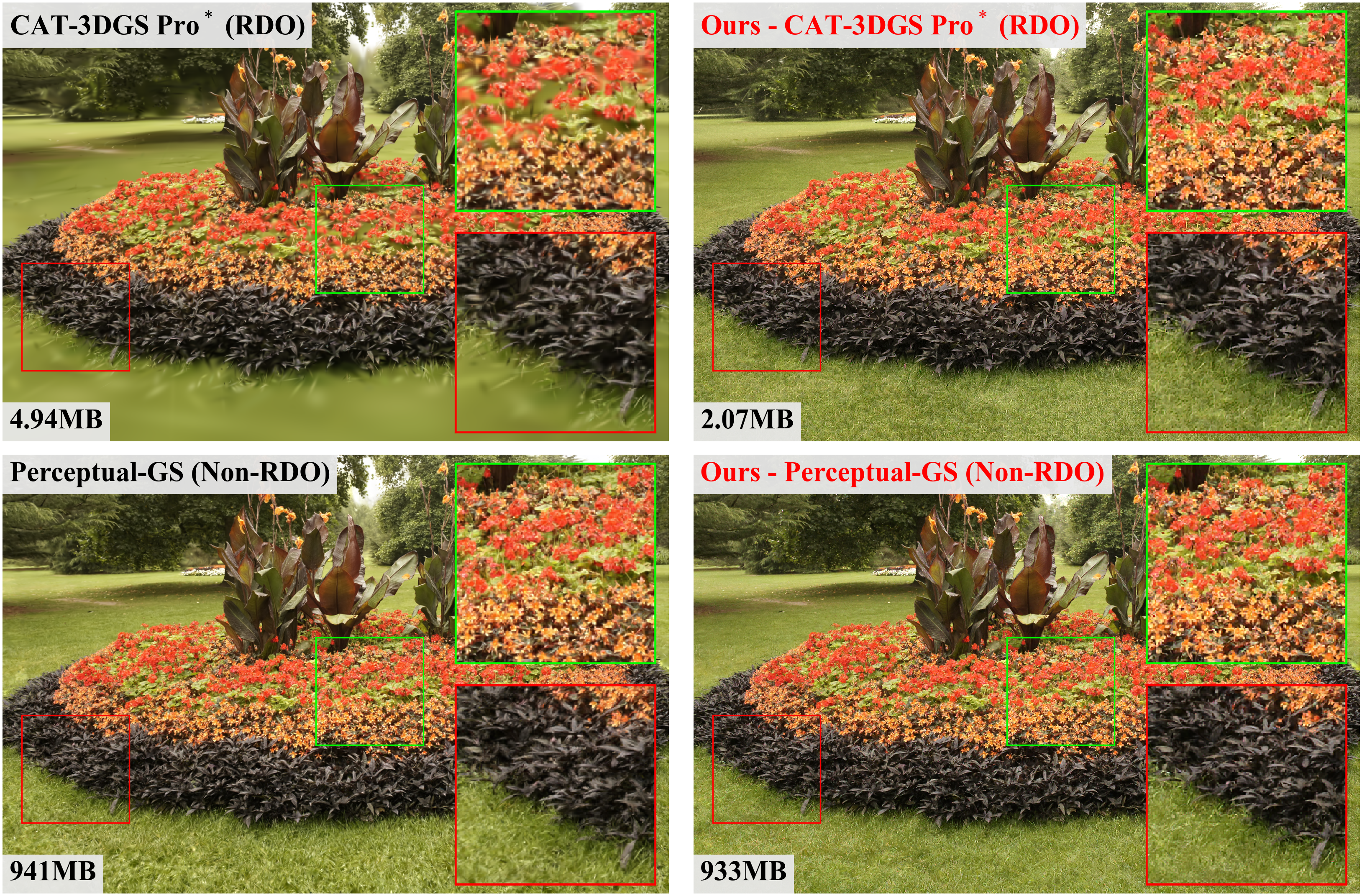}
    \end{subfigure}

    \caption{Comparison of 3DGS  frameworks. \textbf{(a)} Traditional methods often struggle to reproduce high-frequency textures due to suboptimal distortion metrics. \textbf{(b)} Recently, Perceptual-GS~\cite{zhou2025perceptualgs} learns Gaussian primitives by sensitivity guided densification to improve perceptual quality. \textbf{(c)} Our plug-and-play perceptual wrapper integrates a texture synthesis network conditioned on Plücker embeddings and pseudo-random Gaussian noise, supervised by Wasserstein Distortion (WD) to enhance perceptual quality. As demonstrated in (d), our framework consistently improves the perceptual quality on diverse 3DGS variants. The numbers in the bottom-left corner of each image denote the file size in megabytes (MB).}
    \label{fig:teasers}
\end{figure}

3D Gaussian Splatting (3DGS)~\cite{kerbl3Dgaussians} has recently emerged as a promising representation for static 3D scenes, powering immersive applications such as free-viewpoint rendering and scene understanding. Considerable research has been advancing 3DGS representations to achieve (1) high reconstruction quality, (2) compact storage, and (3) fast rendering. Existing approaches include non–rate-distortion-optimized (non-RDO) methods~\cite{scaffoldgs, ren2024octree, sun2024f, zhou2025perceptualgs,mallick2024taming, ren2025fastgs, hanson2025speedy, Zhang2025GaussianSpa,franke2025vr, lin2025metasapiens, ali2024trimming, lee2024compact, fan2024lightgaussian, papantonakis2024reducing, Chen_2025_CVPR, fang2024mini, girish2024eagles, hanson2025pup} and rate-distortion-opti\-mized (RDO) methods~\cite{zhan2025cat, hac++2025, wang2024contextgs, liu2024hemgs,liu2024compgs, wang2024end, zhan2025cat3dgs,hac2024}. The latter distinguishes from the former by integrating additionally entropy coding into the 3DGS training process, often in an end-to-end fashion, for efficient storage and/or transmission. 

However, 3DGS representations can suffer from an excessive number of Gaussian primitives, particularly when modeling scenes with high-frequency textures. This arises because they are optimized primarily based on minimizing a combination of L1 and structural similarity (SSIM) losses~\cite{SSIM} to preserve image fidelity (\cref{fig:teasers}a). In such cases, achieving high reconstruction fidelity typically comes at the expense of increased storage requirements and reduced rendering speed. To date, relatively little research has investigated how the characteristics of human perception can be leveraged to improve 3DGS representations in terms of reconstruction quality, storage size, and rendering speed. As one of the few efforts in this direction, Perceptual-GS~\cite{zhou2025perceptualgs} introduces adaptive densification of Gaussian primitives according to the local perceptual sensitivity of rendered images, assuming that human vision is particularly sensitive to edges in rendered images (\cref{fig:teasers}b). Although demonstrating promising results, it still relies on the conventional mixture of L1 and SSIM losses~\cite{SSIM}, which remains suboptimal for optimizing perceptual quality. Moreover, it does not address how texture regions could be reproduced more effectively by exploiting the reduced sensitivity of peripheral vision in these areas.

Over the years, the image and video compression community has sought to develop objective quality metrics that correlate strongly with human perception, remain computationally efficient, and are ideally differentiable. In parallel, extensive research has utilized texture synthesis techniques as a means of preserving perceived image quality while making the content easier to compress~\cite{filmgrainAV1, ameur2023deepfinegrain}. A notable recent study in~\cite{Balle_2025_CVPR} proposes a multi-scale, local Wasserstein Distortion (WD). Extensive human-rating evaluations confirm its effectiveness both as a perceptual quality measure and as an effective training objective that facilitates texture synthesis through shared common randomness in the context of overfitted image codecs. 

Inspired by~\cite{Balle_2025_CVPR} and recognizing the inefficiency of the 3DGS representation in reproducing highly textured regions, we make the first attempt to incorporate perceptual optimization and texture synthesis into the training and decoding stages of 3DGS. This task presents three unique challenges. \textbf{First}, unlike grid-based image representations, 3DGS representations are inherently \textit{unstructured} and \textit{sparse}. From a holistic perspective, Gaussian primitives resemble point clouds, making it difficult to directly synthesize textures in the 3DGS domain. \textbf{Second}, the requirement of novel view synthesis complicates the decoder design, as it must perform \textit{content-adaptive} texture synthesis on \textit{unseen} views. This differs drastically from the image case~\cite{Balle_2025_CVPR}, where the decoder is overfitted to a given, fixed view. \textbf{Third}, 3DGS representations exist in many different forms with distinctive training procedures, making it desirable to have a \textit{general} framework that can accommodate this diversity.      

Our work introduces a plug‑and‑play perceptual wrapper (\cref{fig:teasers}c) that operates on rendered images from 3DGS representation to improve perceptual quality. It tackles the first challenge by deploying a synthesis network in the image domain--rather than the 3DGS domain--to update the rendered outputs. Its add-on nature allows seamless compatibility with any existing 3DGS representation, thus overcoming the third challenge. Moreover, the synthesis network makes the update content- and view-dependent by leveraging the initial rendered image, per-pixel 6D Plücker embeddings, and pseudo-random Gaussian noise, thereby addressing the second challenge. To minimize its impact on rendering speed, the synthesis network adopts a lightweight implementation. Finally, our perceptual wrapper is trained end-to-end along with the core 3DGS representation using the WD loss to optimize perceptual quality. 

The main contributions of this work are three-fold: (1) it marks the first attempt to integrate image-based texture synthesis and perceptual optimization into the 3DGS training and inference pipelines; (2) it introduces a plug-and-play perceptual wrapper that can be applied to existing 3DGS representations to improve perceptual quality; and (3) it presents extensive human-rating experiments that validate the effectiveness of the proposed approach (\cref{fig:teasers}d).

\begin{figure}[t]
    \centering
    \begin{subfigure}{\linewidth}
        \centering
        \includegraphics[width=\linewidth]{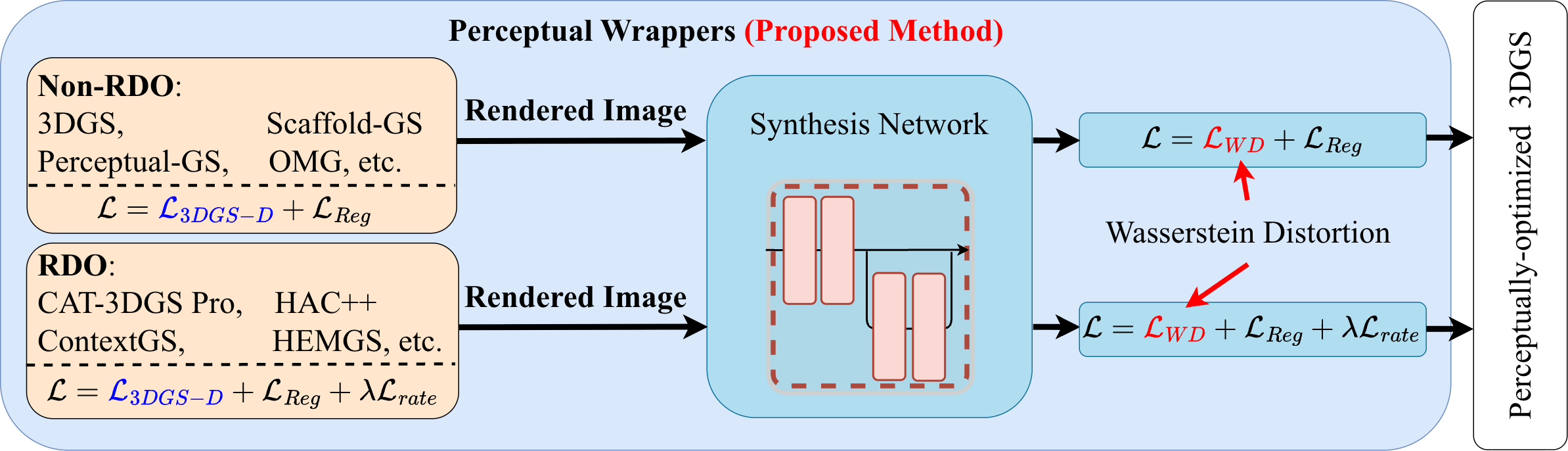}
    \end{subfigure}

    \caption{Overview of 3DGS related works, their training objectives, and their relation to our proposed method.}
    \label{fig:related_work}
\end{figure}
\section{Related Work: 3D Gaussian Splatting}

Existing 3DGS works can be broadly categorized into (i) making 3DGS representations more compact, referred to as non-RDO methods~\cite{scaffoldgs, ren2024octree, sun2024f,mallick2024taming, ren2025fastgs, hanson2025speedy, Zhang2025GaussianSpa,zhou2025perceptualgs,franke2025vr, lin2025metasapiens, ali2024trimming, lee2024compact, fan2024lightgaussian, papantonakis2024reducing, Chen_2025_CVPR, fang2024mini, girish2024eagles, hanson2025pup}, and (ii) compressing them with entropy coding in an RDO manner~\cite{zhan2025cat, hac++2025, wang2024contextgs, liu2024hemgs, liu2024compgs, wang2024end, zhan2025cat3dgs,hac2024}.

Non-RDO 3DGS methods aim to reduce storage and training cost by imposing structure on Gaussian primitives or improving densification and pruning strategies.
Representative approaches that explicitly structure primitives include Scaffold-GS~\cite{scaffoldgs}, Octree-GS~\cite{ren2024octree}, and F-3DGS~\cite{sun2024f}. In parallel, a line of work~\cite{mallick2024taming, ren2025fastgs, hanson2025speedy, Zhang2025GaussianSpa, ali2024trimming, lee2024compact, fan2024lightgaussian, papantonakis2024reducing, Chen_2025_CVPR, fang2024mini, girish2024eagles, hanson2025pup} focuses on more efficient densification or pruning to mitigate an excessive number of Gaussians.
Another research direction focuses on end-to-end RDO compression. State-of-the-art methods include CAT-3DGS Pro~\cite{zhan2025cat}, HAC++~\cite{hac++2025}, ContextGS~\cite{wang2024contextgs}, and HEMGS~\cite{liu2024hemgs}. These approaches jointly optimize reconstruction quality and bit-rates by quantizing Gaussian parameters and performing entropy coding.
Recently, some works incorporate perceptual factors into 3DGS. Perceptual-GS~\cite{zhou2025perceptualgs} introduces perceptual sensitivity-guided densification. VR-oriented works~\cite{franke2025vr, lin2025metasapiens} exploit foveated rendering to reduce computation in peripheral regions, but are fundamentally tied to gaze-dependent pipelines. 

Despite the diversity in 3DGS representation and compression pipelines, most 3DGS variants share a similar training objective: a pixel-wise distortion term (commonly a combination of L1 and SSIM~\cite{SSIM}, referred to as $\mathcal{L}_{3DGS-D}$) together with a method-specific regularization loss $\mathcal{L}_{Reg}$ and optionally a rate term $\mathcal{L}_{rate}$ for RDO methods. However, $\mathcal{L}_{3DGS-D}$ prioritizes pixel-wise signal fidelity, which is inefficient for high-frequency texture regions. Such losses fail to account for the reduced perceptual sensitivity of human peripheral vision, where statistical plausibility matters more than pixel-exact reproduction. This motivates our perceptual wrapper, a perceptual optimization framework applicable to both non-RDO and RDO 3DGS pipelines, as illustrated in~\cref{fig:related_work}.

\section{Method}
\begin{figure*}[t]
\centering
\includegraphics[width=\linewidth]{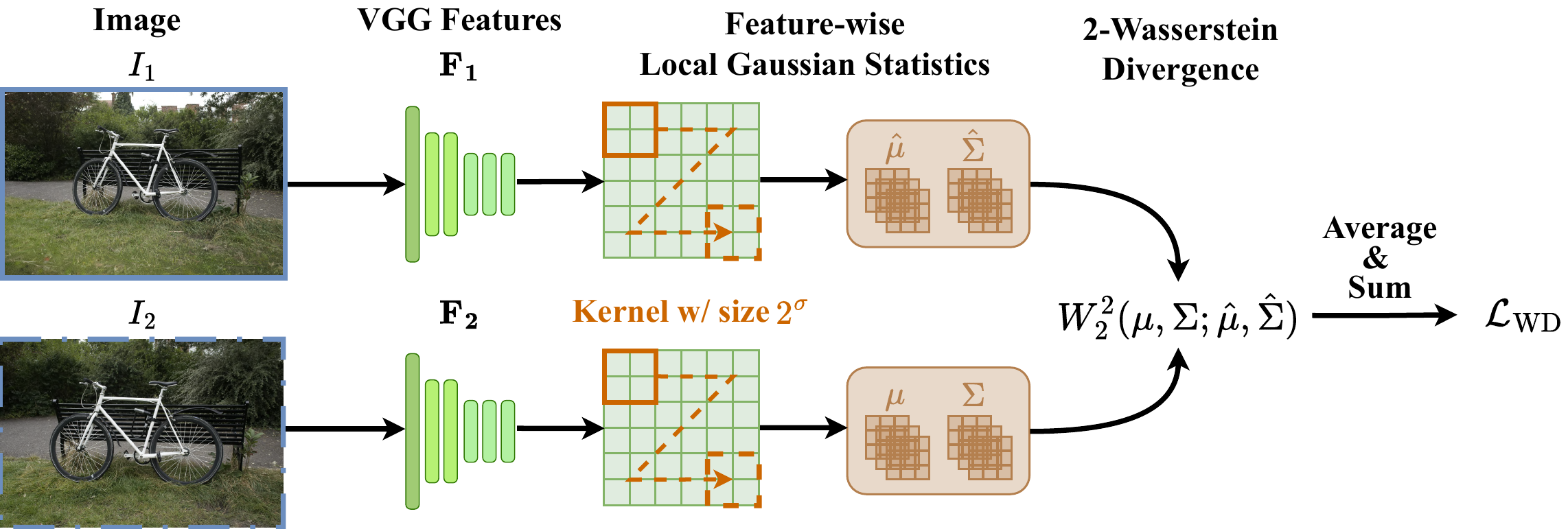}
  \caption{Illustration of Wasserstein Distortion. The distortion is computed as the 2-Wasserstein divergence between pairs of local Gaussian statistics in the feature space.}
  \label{fig:wd}
\end{figure*}
\subsection{Preliminaries}
\subsubsection{Wasserstein Distortion.} Unlike traditional 3DGS methods that minimize pixel-wise objectives, such as L1 and SSIM, our perceptual wrapper adopts \textit{Wasserstein Distortion (WD)}~\cite{qiu2024wasserstein} as the primary training objective. It is crucial to distinguish WD from the well-known \textit{Wasserstein Distance (Earth Mover's Distance)}. The latter quantifies the global divergence between probability distributions (\eg, real images vs. generative outputs), whereas Wasserstein Distortion is an image-to-image metric that focuses on local statistics matching. As shown by Ballé \etal~\cite{Balle_2025_CVPR}, WD provides a closer match to human perceptual judgments than commonly used alternatives such as LPIPS~\cite{lpips} or DISTS~\cite{dists}. 

Similar to LPIPS~\cite{lpips}, WD evaluates distances in a deep feature space using pre-trained VGG~\cite{VGG} embeddings (\cref{fig:wd}). However, it differs fundamentally in how these features are aggregated and compared. To compute WD between two images  $I_1$ and $I_2$, we first extract their multi-scale feature representations $\mathbf{F_1} $ and $\mathbf{F_2}$ from the pre-trained VGG. At each spatial location $(x,y)$ of a feature map, WD applies a pooling kernel of size $\sigma(x,y)$, modeling the surrounding features as a Gaussian distribution. The metric then computes the 2-Wasserstein divergence $W^2_2(\cdot)$ between the corresponding localized Gaussian distributions from the two features. Finally, these local WD values are spatially averaged and summed across feature maps. 
 
 In particular, WD incorporates a user-defined kernel size to balance point-wise and region-wise statistical matching. Given the hyperparameter $\sigma$, the kernel size is defined as $2^\sigma$ and governs the perceptual behavior of the rendered images. As $\sigma \to 0$, WD reduces to a point-wise feature distance that is conceptually similar to LPIPS. This strictly penalizes point-wise deviations between the corresponding feature maps of the two images, approximating human fovea vision (\ie, vision at the center of gaze). Conversely, a larger $\sigma$ encourages texture re-sampling by prioritizing neighborhood-level statistical alignment over point-wise fidelity. This mimics human peripheral vision, allowing the model to synthesize perceptually plausible textures without requiring exact signal reconstruction. By choosing $\sigma$ properly, WD achieves a trade-off between perceptual quality and reconstruction fidelity. Furthermore, a spatially varying kernel size can be adopted to offer greater flexibility for perceptual optimization~\cite{Balle_2025_CVPR}.

\subsubsection{Overfitted Image Codecs with Common Randomness.} 
To showcase the effectiveness of WD, Ballé \etal~\cite{Balle_2025_CVPR} train a per-image overfitted image codec using WD as the distortion objective. An overfitted image codec typically consists of grid-based learnable image latents, a hyperprior network, and a lightweight decoder. The hyperprior network models the probability distribution of the image latents for entropy coding, while the lightweight decoder reconstructs RGB samples in a sample-by-sample manner by decoding the latents. All three components are overfitted to the input image by back-propagating a training loss composed of the rate and distortion terms. In~\cite{Balle_2025_CVPR}, conventional distortion metrics, such as L1, MSE, SSIM~\cite{SSIM}, or LPIPS~\cite{lpips}, are replaced with WD to enable perceptual optimization. Specifically, they adopt a spatially varying $\sigma$ kernel to allocate more coding bits to salient regions, and supply the decoder with additional \textit{pseudo-random Gaussian noise} to facilitate texture synthesis. This Gaussian noise, shared between the encoding and decoding processes by synchronizing a random seed, is known as \textit{common randomness}. Introducing this additional randomness at the decoder input proves effective in synthesizing perceptually plausible patterns in textured regions.

\begin{figure*}[t]
\centering
\includegraphics[width=\linewidth]{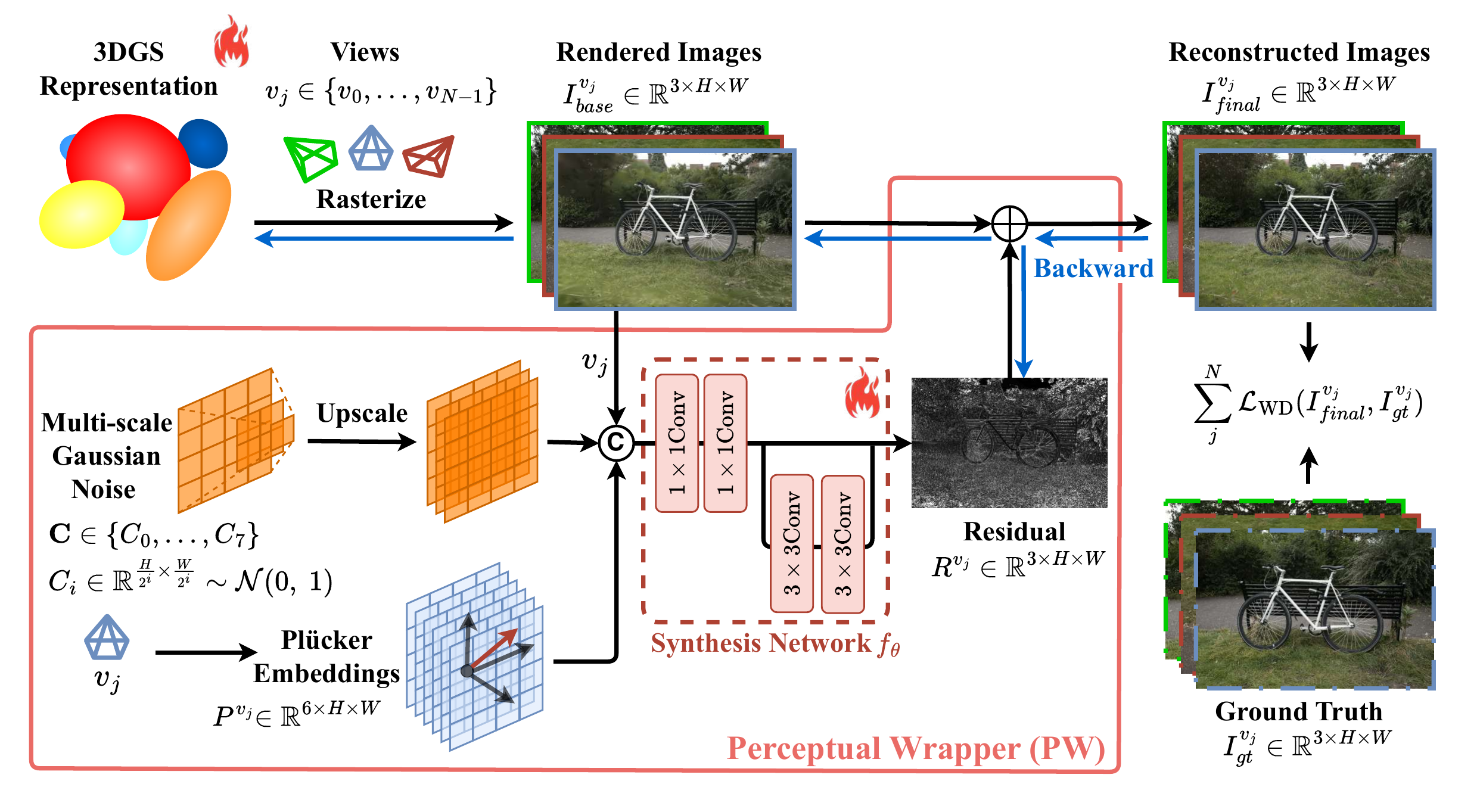}
  \caption{Overview of the proposed framework. The perceptual wrapper uses a synthesis network to refine the rendered image $I_{base}^{v_j}$ in a content- and view-dependent manner. The whole pipeline is optimized via minimizing a Wasserstein Distortion loss $\mathcal{L}_{WD}$ across training views.}
  \label{fig:overview}
\end{figure*}

\subsection{Perceptual Wrapper for 3DGS}

\subsubsection{Overview.} This work seeks to integrate perceptual optimization and texture synthesis into both the training and decoding stages of per-scene 3DGS representations, addressing their inefficiency in reproducing highly textured regions. Under strict storage or transmission bandwidth constraints, these regions often exhibit noticeable blurry artifacts. While texture synthesis techniques for perceptual optimization under representation cost constraints have been extensively studied in the image and video domains, their application to 3DGS remains largely unexplored. 

Per-scene 3DGS representations parallel overfitted image codecs in that both approaches overfit the representations and the decoder to a specific instance. However, 3DGS representations differ fundamentally from grid-based image latents. They are inherently sparse and unstructured, making texture synthesis directly in the 3DGS domain highly challenging. The requirement of novel view synthesis further demands that texture synthesis adapt to viewpoint. Moreover, 3DGS representations continue to evolve rapidly across applications. There is significant value in a solution that can generalize across 3DGS variants. 

Recognizing these differences and requirements, we propose a versatile, plug-and-play perceptual wrapper (PW), as shown in Fig.\ref{fig:overview}. By directly operating on the 2D rendered image rather than the 3D Gaussian primitives, our approach avoids the complexities of unstructured texture generation in the 3DGS domain and integrates seamlessly with any existing 3DGS representation. Our perceptual wrapper comprises two primary components: a multi-scale, 2D pseudo-random Gaussian noise and a lightweight synthesis network. Alongside the proposed module, we introduce a training strategy that replaces the traditional L1 or SSIM~\cite{SSIM} loss with WD, explicitly optimizing the model to better align with human perceptual preferences.

\subsubsection{View-Consistent Pseudo-Random Gaussian Noise.} To facilitate texture synthesis in the image domain, the input to our lightweight decoder includes the rendered image $I^{v_j}_{base} \in \mathbb{R}^{3 \times H \times W}$ from a specific view $v_j \in \{v_0, \dots, v_{N-1}\}$ and a multi-scale pseudo-random Gaussian noise pyramid $\mathbf{C} = \{C_i\}_{i=0}^7$. This noise is structured as an 8-level spatial pyramid, modeling texture signals across various frequencies, where $C_i \in \mathbb{R}^{\frac{H}{2^i} \times \frac{W}{2^i}}$. The highest-resolution level ($C_0$) matches the spatial dimensions of $I^{v_j}_{base}$, while subsequent levels are recursively downsampled by a factor of two. During the decoding phase, noise at each resolution level is bilinearly upsampled to the original image resolution. Each grid point value in $\mathbf{C}$ is sampled independently from a standard normal distribution $\mathcal{N}(0, 1)$.
This noise, which serves as a common source of randomness, remains fixed across views, and between the encoding and decoding processes. A fixed random seed is employed during training and decoding to eliminate the overhead of signaling the noise. Our view-consistent strategy ensures that the same noise can be applied to any novel view at inference time.

\subsubsection{Synthesis Network.} Our synthesis network is designed to generate perceptually plausible texture details that are not captured by the underlying 3DGS representation. It operates in a content- and view-dependent manner. The network adopts an architecture similar to the COOL-CHIC family~\cite{ladune2023coolchic, kim2024c3}, consisting of two $1 \times 1$ convolutional layers, followed by two $3 \times 3$ residual convolutional layers. Its lightweight design ensures that the module incurs negligible transmission and memory overhead while maintaining \textcolor{red}{real-time} rendering speeds. In addition to the rendered image (which makes the network output content-dependent) and the multi-scale Gaussian noise, the network also takes as input the Plücker embeddings $P^{v_j}\in\mathbb{R}^{6\times H\times W}$ of the target view $v_j$, which serve as a conditioning signal for decoding. The Plücker embedding~\cite{NEURIPS2021_a11ce019} for a pixo at coordinates $(x,y)$  in a rendered image is a 6D vector $P(x,y) = (\mathbf{d},\mathbf{o} \times \mathbf{d})$, with $\mathbf{d}$ being the normalized viewing direction:

\begin{equation}
  \mathbf{d} = \frac{R K^{-1}[x,y,1]^T}{\|RK^{-1}[x,y, 1]^T\|} \in \mathbb{R}^3, 
\end{equation}
where $K \in \mathbb{R}^{3\times3}$ is the camera intrinsic matrix and $R \in \mathbb{R}^{3\times3}$ is the camera-to-world rotation matrix, $\mathbf{o} \in \mathbb{R}^3$ is the camera center in world coordinates, and $\times$ denotes cross product. By conditioning on this view-dependent per-pixel ray information, the network can distinguish rendered pixels with similar appearance but different viewing rays, and learns to synthesize view-dependent high-frequency details at geometrically grounded locations. Our ablation study (Section~\ref{sec:plucker_ablation}) confirms its effectiveness as a crucial signal that enables the synthesis network to adapt its behavior to different viewpoints.

After initial rendering, for a specific view $v_j$, the base image $I_{base}^{v_j}$ is concatenated with the Gaussian noise $\mathbf{C}$ along with the Plücker embeddings $P^{v_j}$. This combined input is processed through the synthesis network $f_\theta$, which generates perceptually plausible texture details:
\begin{equation}
  R^{v_j} = f_{\theta}(\{I_{base}^{v_j},\mathbf{C},P^{v_j}\})\in \mathbb{R}^{3\times H\times W}. 
\end{equation}
As a residual signal, these high-frequency details are added to the rendered image to arrive at the final output:
\begin{equation}
  I_{final}^{v_j} = I_{base}^{v_j}+R^{v_j}. 
\end{equation}

\subsubsection{Per-scene Optimized Training.} 
We adopt a two-stage, per-scene optimized training procedure. The learning objectives include the 3DGS representation and the synthesis network. 
We note that directly optimizing the full pipeline with WD from the outset can destabilize the 3DGS training, particularly when Gaussian densification/pruning are still evolving and the representation has not yet converged to a stable geometry.

In the first stage, we warm up the underlying 3DGS representation using its original training objective $\mathcal{L}_{\text{3DGS-D}}$, together with the method-specific regularization term $\mathcal{L}_{\text{Reg}}$ and, for RD-optimized methods, the rate term $\mathcal{L}_{\text{rate}}$ when applicable. This stage ensures stable densification and pruning of Gaussian primitives, yielding a well-formed 3DGS representation. It lasts for 15,000 iterations, during which the Gaussians are actively refined via densification/pruning.

In the second stage, we replace the distortion term $\mathcal{L}_{\text{3DGS-D}}$ with the WD $\mathcal{L}_{\text{WD}}$ while retaining the other terms from the first stage. At this point, the synthesis network is activated and jointly optimized with the 3DGS representation. Beginning at iteration 15,000, the WD between the reconstructed image $I_{final}^{v_j}$ and the ground truth $I_{gt}^{v_j}$ is evaluated and averaged across training views:
\begin{equation}
\mathcal{L} = \mathcal{L}_{\text{WD}}(I_{\text{final}}^{v_j}, I_{\text{gt}}^{v_j}) \;+\; \mathcal{L}_{\text{Reg}} \;+\; \lambda \mathcal{L}_{\text{rate}},
\end{equation}
where the rate term $\mathcal{L}_{\text{rate}}$ is included only for RD-optimized methods.

\section{Experimental Results}
We implement all model variants in PyTorch
and perform training on a single NVIDIA Tesla V100 GPU. Training details and hyperparameters are provided in the supplementary material for reproducibility.

\subsection{Implementation Details}
\subsubsection{Baselines.}
To validate the generality of our method across different 3DGS representations, we augment several representative 3DGS representations with our perceptual wrapper (PW) and compare them to the original counterparts. We consider both RDO (\eg, CAT-3DGS Pro~\cite{zhan2025cat} and HAC++~\cite{hac++2025}) and non-RDO (\eg, 3DGS~\cite{kerbl3Dgaussians}, OMG~\cite{lee2025optimized} and Per\-ceptual-GS~\cite{zhou2025perceptualgs}) methods. 

For RDO methods, we assess reconstructions decoded from compressed bitstream at three dataset-average rate points, ensuring that the baselines and our augmented variants operate under similar bit-rates. For CAT-3DGS Pro, we adopt an enhanced variant (denoted CAT-3DGS Pro$^*$), which adopts the INR-based hyperprior from CSGaussian~\cite{Tseng_2026_WACV}.

For non-RDO methods, aligning their model sizes for rate matching becomes difficult without altering their default training configurations or model capacity. In our experiments, we adopt their released code directly. OMG provides three variants (OMG-XS/M/XL) of different model sizes. These variants share the same architecture, but differ in pruning strength.

\subsubsection{Datasets.} We conduct experiments on the Mip-NeRF 360 dataset~\cite{barron2022mip}, which contains five outdoor scenes (\textit{bicycle, flowers, garden, stump, treehill}) and four indoor scenes (\textit{bonsai, counter, kitchen, room}), all evaluated at a resolution of $1600 \times 1063$. The choice of this dataset for evaluation is constrained by the resources and time required to perform human ratings. Additional results on Tanks \& Temples ~\cite{10.1145/3072959.3073599} and Deep Blending~\cite{10.1145/3272127.3275084} datasets are provided in the supplementary material.

\subsubsection{Human Rating for Subjective Quality Evaluation.} We compare the rate-distortion performance of competing methods by visualizing their rate-distortion curves. The horizontal axis represents the rate, defined as the compressed bitstream size for RDO methods and the model size for non-RDO methods. In both cases, the size accounts for the 3DGS representation and the synthesis network. The vertical axis denotes reconstruction quality, measured by Elo score derived from human ratings. Elo scoring is extensively used in Challenge on Learned Image Compression (CLIC)~\cite{CLIC}. To collect Elo scores, we follow the CLIC 2025 test protocol and conduct human rating on the Mabyduck platform~\cite{mabyduck}, which implements  fair subjective evaluation environment with the Bayesian Elo rating system~\cite{bayesian_elo}. During evaluation, ground-truth images are displayed on the left for reference, while two reconstructed versions, corresponding to two different methods, are shown on the right, allowing raters to toggle between them with a key press. All images are randomly cropped to a resolution of $512 \times 512$, and raters may request additional random crops to examine different spatial regions. They are asked to select the reconstruction that appears perceptually closer to the ground-truth image. For interpretation, a higher Elo score indicates a reconstruction that is perceptually closer to the reference. If the Elo difference between two methods is 0, they each have a 50\% win probability (equally matched). A difference of 100 points corresponds to a 64\% win probability, 200 points to 76\%, and 400 points to 91\%.

In total, we collect 6500 pairwise comparisons from 65 raters, with each rater evaluating 100 image pairs. Prior to evaluation, raters were informed of the study objectives and completed a pre-screening session to familiarize themselves with the evaluation procedure. The pre-screening session also allows us to filter out unreliable raters since all test questions served as golden questions, in which one of the reconstructions is replaced with the ground-truth reference image. Note that our evaluation is restricted to \textit{intra-method} comparisons. For each baseline method, human raters compare only the baseline version against its enhanced counterpart using our perceptual wrapper. Consequently, the reported Elo scores reflect relative improvements for each baseline method and should not be used for inter-method comparisons (\eg, comparing enhanced versions of different methods), as no direct pairwise preference data were collected across different methods. In addition to human evaluation, we also report results with widely used objective quality metrics, including LPIPS~\cite{lpips}, DISTS~\cite{dists}, PSNR, and SSIM~\cite{SSIM}.

\begin{figure}[t]
    \centering
    \begin{subfigure}[t]{0.64\linewidth}
        \centering
        \begin{subfigure}[t]{0.49\linewidth}
            \centering
            \includegraphics[width=\linewidth]{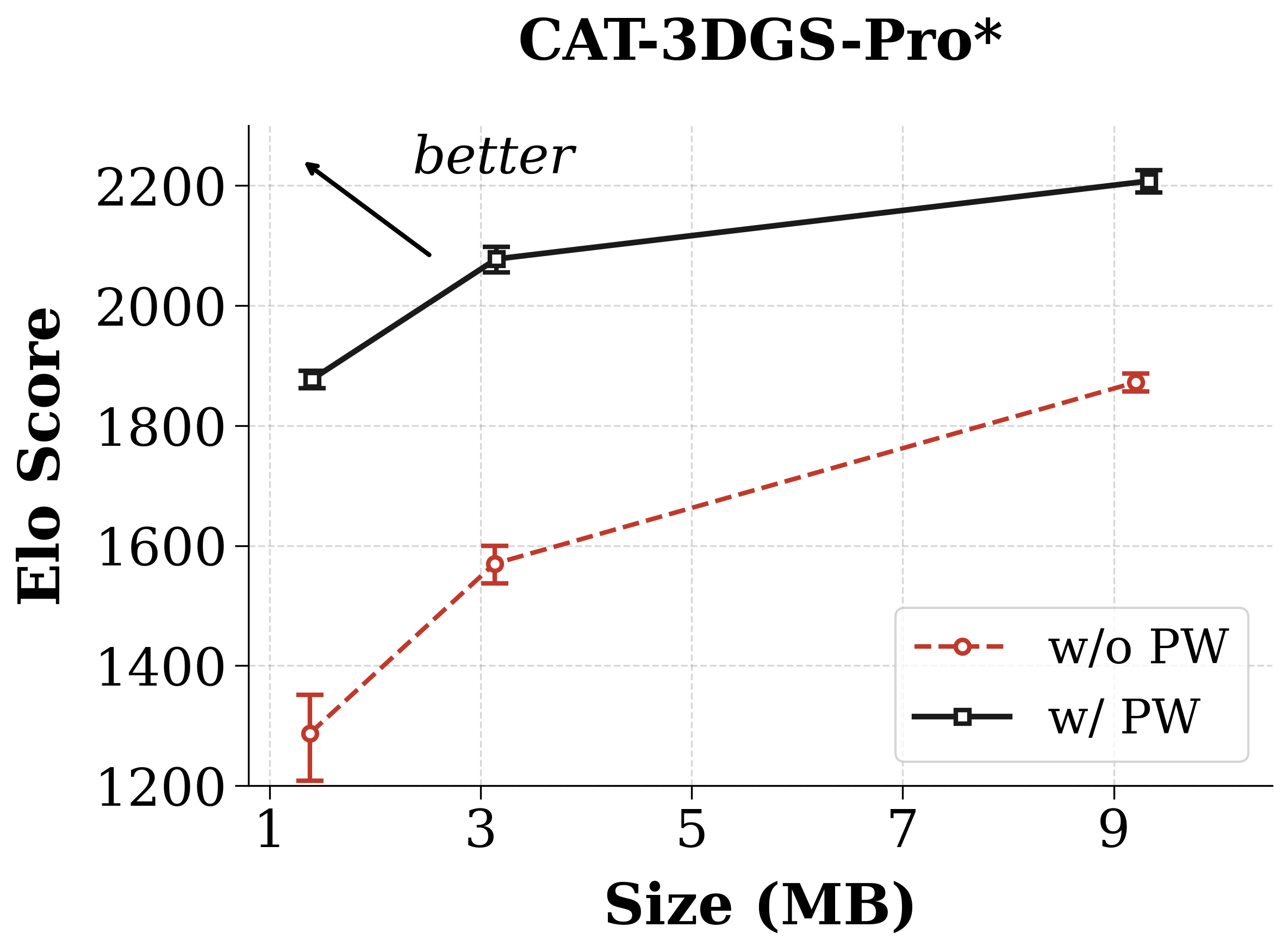}
        \end{subfigure}
        \hfill
        \begin{subfigure}[t]{0.49\linewidth}
            \centering
            \includegraphics[width=\linewidth]{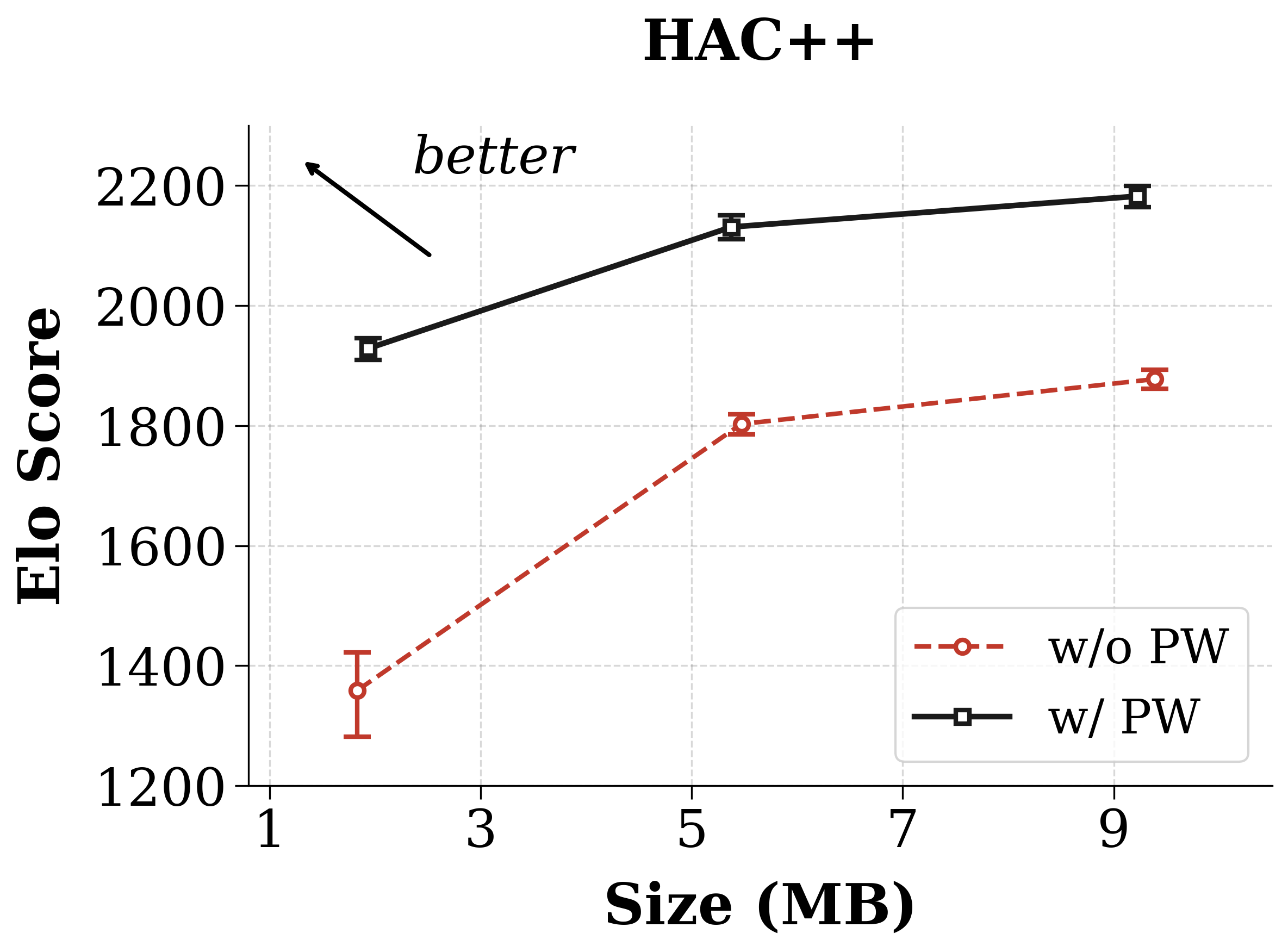}
        \end{subfigure}
        \caption{RDO methods: CAT-3DGS Pro$^*$~\cite{zhan2025cat} and HAC++~\cite{hac++2025}.}
        \label{fig:rdo_curves_rdo}
    \end{subfigure}
    \hfill
    \begin{subfigure}[t]{0.35\linewidth}
        \centering
        \includegraphics[width=0.896
        \linewidth]{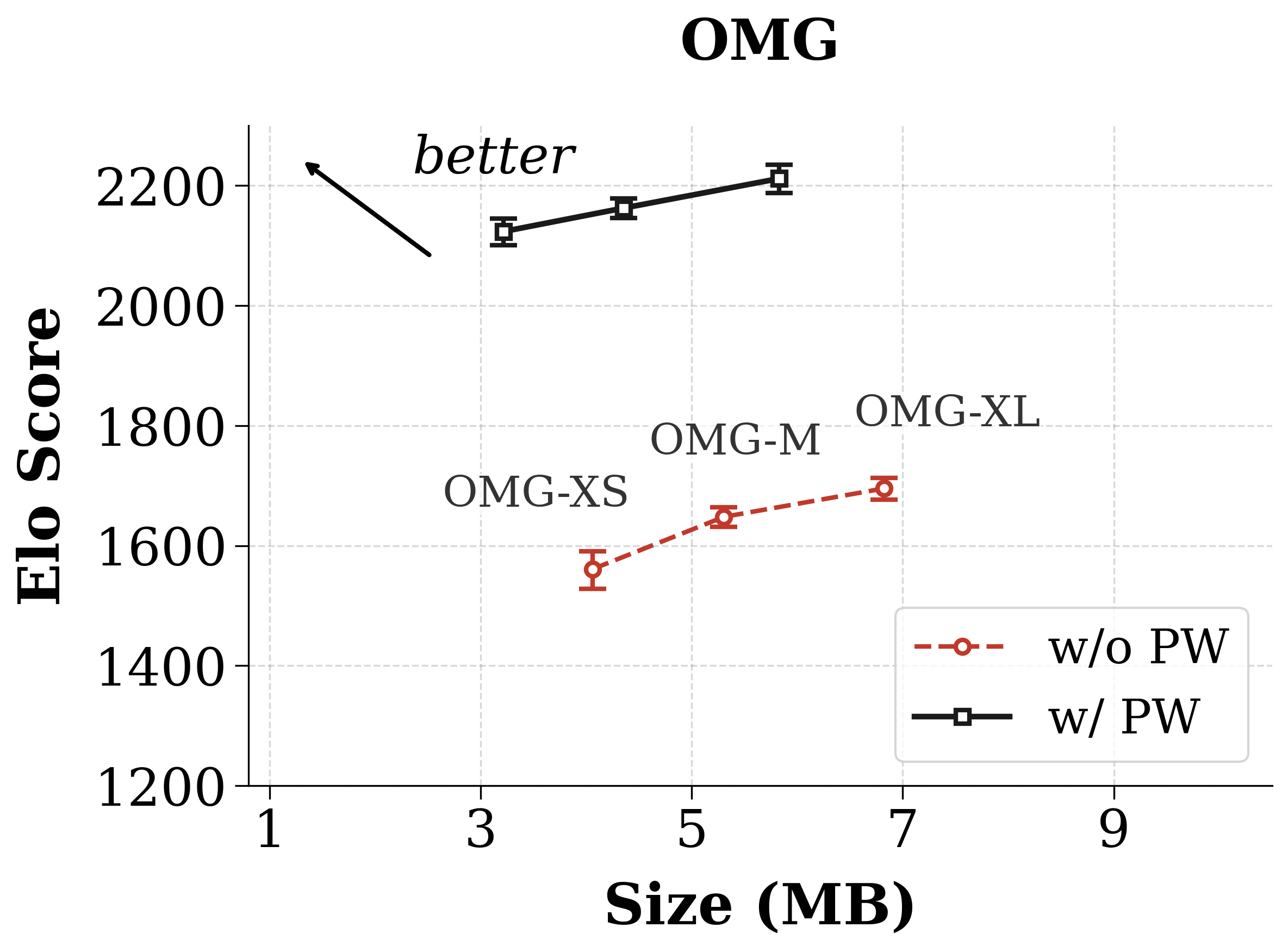}
        \captionsetup{justification=raggedright,singlelinecheck=false}
        \caption{Non-RDO method: OMG~\cite{lee2025optimized}.}
        \label{fig:rdo_curves_nrdo}
    \end{subfigure}
    \caption{Rate-distortion performance across RDO and non-RDO methods
.}
    \label{fig:rdo_curves}
\end{figure}
\begin{figure*}[t]
\centering
\includegraphics[width=\linewidth]{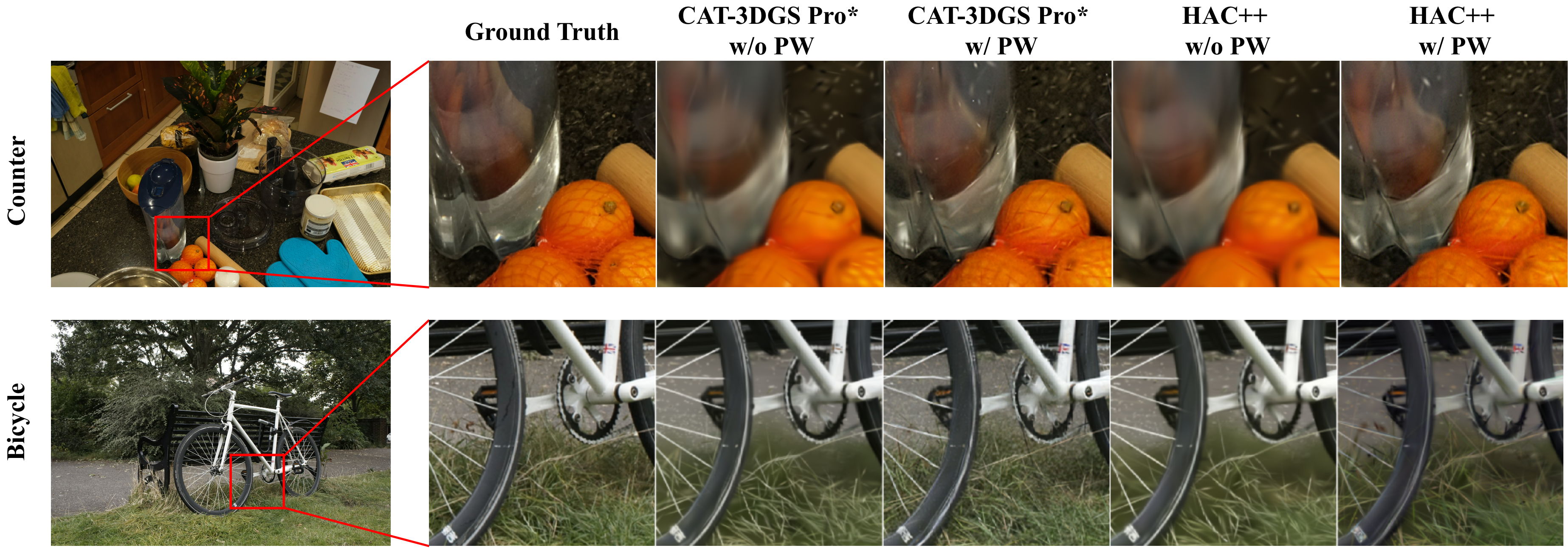}
\caption{Qualitative results of our perceptual wrapper (PW) integrated into RDO methods, CAT-3DGS Pro$^*$~\cite{zhan2025cat} and HAC++~\cite{hac++2025}.}
  \label{fig:rdo_visualization}
\end{figure*}
\begin{table}[t]
    \centering
    \setlength{\tabcolsep}{8pt}
    \caption{Quantitative comparison of non-RDO methods with and without our perceptual wrapper (PW).}
    \label{tab:nrdo_comparison}
    \begin{adjustbox}{max width=\linewidth}
    \begin{tabular}{lcccccc}
    \toprule
    Method & File size (MB) & Elo score $\uparrow$ & PSNR $\uparrow$ & SSIM $\uparrow$ & LPIPS $\downarrow$ & DISTS $\downarrow$ \\
    \midrule
    3DGS~\cite{kerbl3Dgaussians}      &  623      &  1807  &  \textbf{27.53} & \textbf{0.8131}  &  	0.2207  &  0.0851  \\
    3DGS w/ PW    &  623 &  \textbf{2100}  &  25.93 &0.7231 &  \textbf{0.1945}  &  \textbf{0.0464}  \\
    \specialrule{0.6pt}{0pt}{0pt} 
    Perceptual-GS~\cite{zhou2025perceptualgs}      &  698   &  1899  &  \textbf{27.77} & \textbf{0.8284} &  \textbf{0.1860}  &  0.0498  \\
    Perceptual-GS w/ PW     &  700  &  \textbf{2075}  & 25.68&  0.6990  &  0.1972  &  \textbf{0.0438}  \\
    \bottomrule
    \end{tabular}
    \end{adjustbox}
\end{table}

\begin{figure*}[t]
\centering
\includegraphics[width=\linewidth]{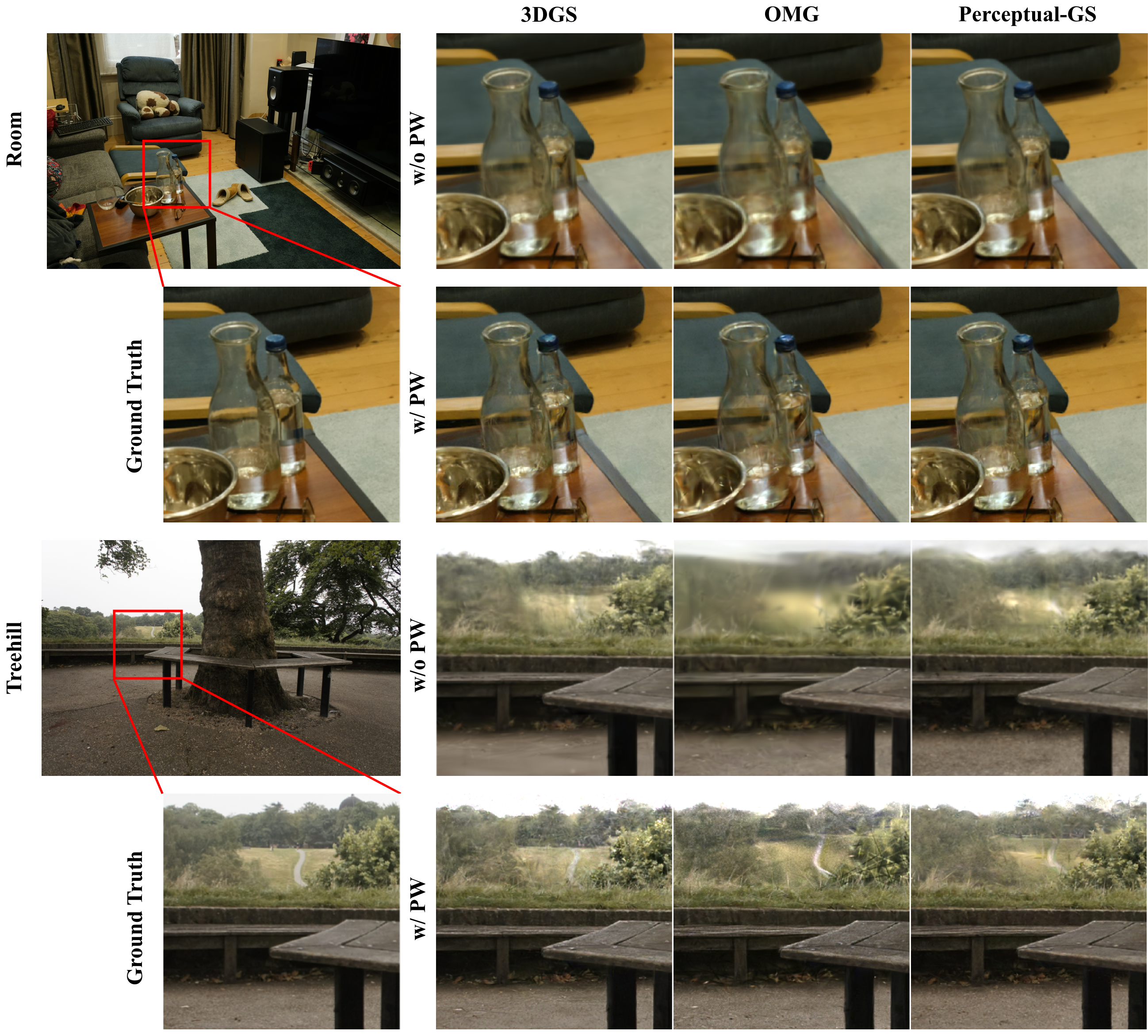}
\caption{Qualitative results of our perceptual wrapper (PW) integrated into non-RDO methods, 3DGS~\cite{kerbl3Dgaussians}, OMG~\cite{lee2025optimized} and Perceptual-GS~\cite{zhou2025perceptualgs}.}
  \label{fig:non_rdo_visualization}
\end{figure*}

\subsection{Quantitative and Qualitative Results}
\label{sect:Q_results}
\subsubsection{Comparisons with RDO-based 3DGS.}
\cref{fig:rdo_curves_rdo} reports the results of our subjective study after integrating our perceptual wrapper into RDO-based 3DGS methods, including CAT-3DGS Pro$^*$ and HAC++. Across all evaluated rate points and baseline methods, our approach yields substantial improvements in perceptual quality at comparable bit-rates. Remarkably, the lowest-rate point achieved by our method attains an Elo score on par with the highest-rate point of the baseline methods (CAT-3DGS Pro$^*$ and HAC++), corresponding to a $6\times$ reduction in bit-rate at equivalent perceptual quality. Additional objective results (PSNR, SSIM, LPIPS, DISTS) are provided in the supplementary material.

\cref{fig:rdo_visualization} presents qualitative comparisons at the lowest rate point for a novel view. Under stringent rate budgets, RDO-based 3DGS methods typically blur fine texture details to conserve coding bits. In contrast, at comparable bit-rates, our perceptual wrapper exploits pseudo-random Gaussian noise to synthesize perceptually convincing texture details. More results are presented in the supplementary material.

\subsubsection{Comparisons with Non-RDO 3DGS.}
\cref{fig:rdo_curves_nrdo} reports results for OMG~\cite{lee2025optimized}, a non-RDO 3DGS method. The same trend observed in RDO-based methods holds here: our method consistently improves Elo scores at all operating points. Notably, applying our perceptual wrapper to the smallest OMG variant (\ie OMG-XS) yields a higher Elo score than the original OMG-XL without perceptual wrapper, indicating that our method can deliver superior perceptual quality at a substantially reduced model size.

For 3DGS~\cite{kerbl3Dgaussians} and Perceptual-GS~\cite{zhou2025perceptualgs}, only a single model size is available from their released code. As shown in \cref{tab:nrdo_comparison}, integrating our perceptual wrapper has little impact on model size while consistently yielding higher Elo scores. It is noteworthy that our perceptual wrapper also delivers meaningful Elo score gains when applied to Perceptual-GS, which already incorporates perceptual sensitivity into its densification strategy. Qualitative comparisons in \cref{fig:non_rdo_visualization} further confirm its benefits for perceptual quality. This suggests that our approach is complementary to Perceptual-GS, addressing texture synthesis challenges that sensitivity-guided densification alone cannot fully resolve. As expected, we observe a trade-off between pixel-level fidelity (PSNR, SSIM) and perceptual quality (Elo score, LPIPS, DISTS).

\subsection{Effect of Plücker Embeddings}
\label{sec:plucker_ablation}
\begin{figure}[tb]
\centering
\includegraphics[width=\linewidth]{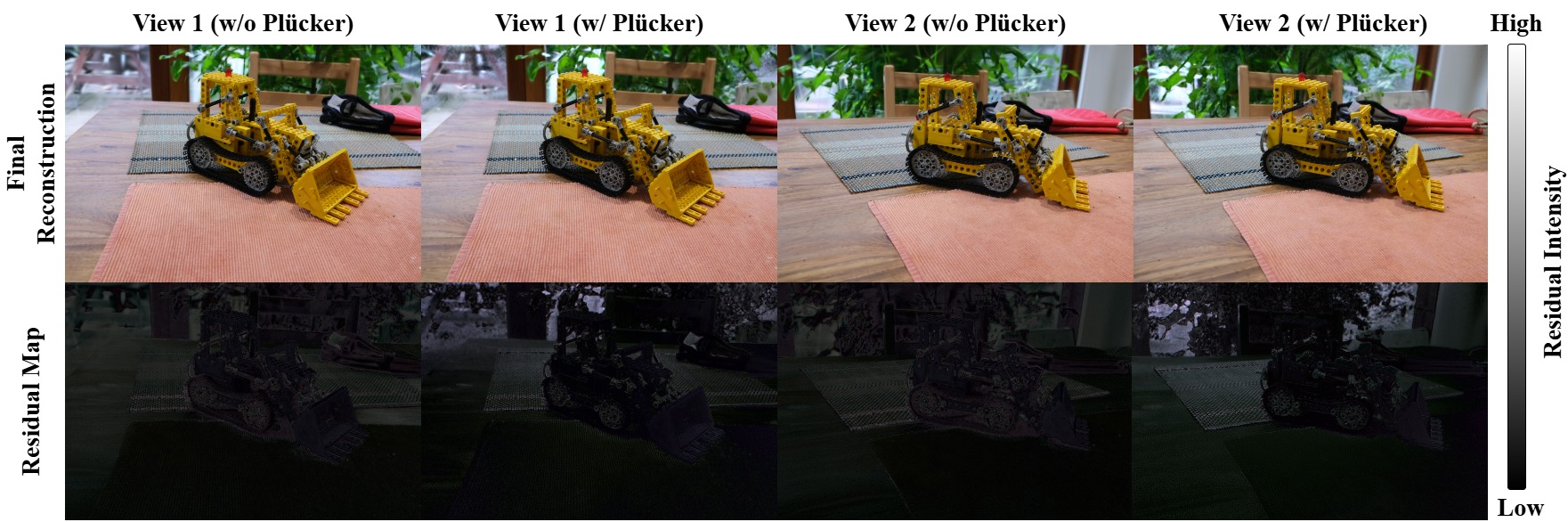}
\caption{Comparison of residual maps $R^{v_j}$ (the second row) with and without Plücker embeddings. The first row shows their corresponding final outputs.}
  \label{fig:plucker}
\end{figure}

We ablate the effect of Plücker embeddings in the synthesis network. \cref{fig:plucker} visualizes the synthesized residuals with and without Plücker embeddings across different viewpoints, based on CAT-3DGS Pro$^*$. Without Plücker embeddings, the residual is spatially diffused and uniformly low in magnitude, indicating that the network applies only weak texture updates across the entire image. In contrast, Plücker embeddings enable the network to concentrate residuals in regions where Gaussian primitives struggle to faithfully reproduce fine textures such as the background foliage in the \textit{kitchen} scene across different viewpoints. This confirms that the network learns to exploit geometric cues to target perceptually challenging regions.

\subsection{3DGS Visualizations}
\begin{figure*}[t]
\centering
\includegraphics[width=\linewidth]{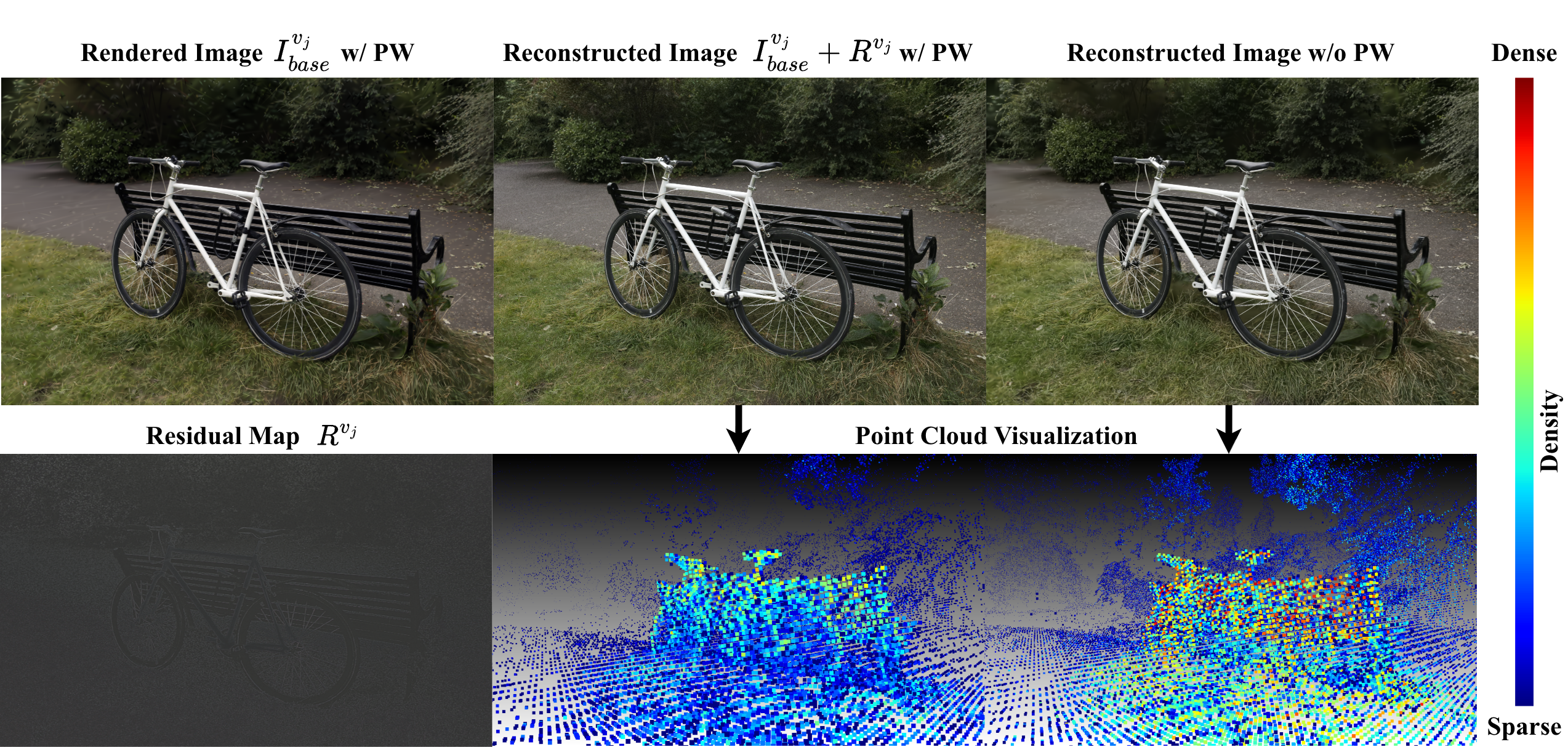}
\caption{Visualization of 3DGS point clouds with and without our perceptual wrapper.}
  \label{fig:point_cloud}
\end{figure*}

\cref{fig:point_cloud} visualizes the Gaussian primitive density maps for two variants (w/ and w/o our perceptual wrapper (PW)) of CAT-3DGS Pro$^*$ at comparable Elo scores. Brighter regions indicate denser Gaussian concentrations. 

Without our perceptual wrapper, the underlying 3DGS representation allocates a large number of Gaussian primitives to texture-rich regions such as grass. In contrast, with our perceptual wrapper, the model exhibits a significant reduction in Gaussian density in these regions. This confirms that by offloading the reconstruction of high-frequency details to our synthesis network, our perceptual wrapper relieves the 3DGS representation of the burden of explicitly modeling fine-scale textures, enabling a sparser and more efficient set of primitives at equivalent perceptual quality.

\section{Discussion} This work represents the first attempt to introduce texture synthesis and perceptual optimization into the 3DGS pipeline. While our perceptual wrapper demonstrates promising improvements in perceptual quality and integrates seamlessly with diverse 3DGS representations, it inevitably incurs additional training and inference time. On the Mip-NeRF 360 dataset, using CAT-3DGS Pro$^*$ at comparable Elo scores, our perceptual wrapper increases the average per-scene training time from 85 to 304 minutes due to the extra computation required for WD evaluation, which involves multi-scale VGG feature extraction at each iteration. Likewise, the average rendering speed decreases from 66 FPS to 31 FPS, as the synthesis network introduces an additional forward pass for each rendered image. We consider these costs as a reasonable trade-off given that the compressed file size with perceptual wrapper is reduced by a factor of 6, an 83\% bit-rate saving (\cref{fig:rdo_curves_rdo}), with similar savings observed for HAC++.
The training overhead primarily stems from VGG feature extraction, while the rendering overhead is largely determined by the choice of the synthesis network architecture. Both aspects leave ample room for future improvement.

\bibliographystyle{splncs04}
\bibliography{main}

\begin{thebibliography}{10}
\providecommand{\url}[1]{\texttt{#1}}
\providecommand{\urlprefix}{URL }
\providecommand{\doi}[1]{https://doi.org/#1}

\bibitem{CLIC}
7th challenge on learned image compression ({CLIC} 2025). \url{https://clic2025.compression.cc/} (2025), last accessed 2026/03/04

\bibitem{ali2024trimming}
Ali, M.S., Qamar, M., Bae, S.H., Tartaglione, E.: Trimming the fat: Efficient compression of 3d gaussian splats through pruning. arXiv preprint arXiv:2406.18214  (2024)

\bibitem{ameur2023deepfinegrain}
Ameur, Z., Hamidouche, W., François, E., Radosavljević, M., Menard, D., Demarty, C.H.: Deep-based film grain removal and synthesis. IEEE Transactions on Image Processing  \textbf{32},  5046--5059 (2023). \doi{10.1109/TIP.2023.3308726}

\bibitem{Balle_2025_CVPR}
Ballé, J., Versari, L., Dupont, E., Kim, H., Bauer, M.: Good, cheap, and fast: Overfitted image compression with wasserstein distortion. In: 2025 IEEE/CVF Conference on Computer Vision and Pattern Recognition (CVPR). pp. 23259--23268 (2025). \doi{10.1109/CVPR52734.2025.02166}

\bibitem{barron2022mip}
Barron, J.T., Mildenhall, B., Verbin, D., Srinivasan, P.P., Hedman, P.: Mip-nerf 360: Unbounded anti-aliased neural radiance fields. In: 2022 IEEE/CVF Conference on Computer Vision and Pattern Recognition (CVPR). pp. 5460--5469 (2022). \doi{10.1109/CVPR52688.2022.00539}

\bibitem{bayesian_elo}
Caron, F., Doucet, A.: Efficient bayesian inference for generalized bradley–terry models. Journal of Computational and Graphical Statistics  \textbf{21}(1),  174--196 (2010)

\bibitem{hac2024}
Chen, Y., Wu, Q., Lin, W., Harandi, M., Cai, J.: Hac: Hash-grid assisted context for 3d gaussian splatting compression. In: European Conference on Computer Vision (2024)

\bibitem{hac++2025}
Chen, Y., Wu, Q., Lin, W., Harandi, M., Cai, J.: Hac++: Towards 100x compression of 3d gaussian splatting. IEEE Transactions on Pattern Analysis and Machine Intelligence  \textbf{47}(11),  10210--10226 (2025). \doi{10.1109/TPAMI.2025.3594066}

\bibitem{Chen_2025_CVPR}
Chen, Y., Jiang, J., Jiang, K., Tang, X., Li, Z., Liu, X., Nie, Y.: Dashgaussian: Optimizing 3d gaussian splatting in 200 seconds. In: Proceedings of the IEEE/CVF Conference on Computer Vision and Pattern Recognition (CVPR). pp. 11146--11155 (June 2025)

\bibitem{dists}
Ding, K., Ma, K., Wang, S., Simoncelli, E.P.: Image quality assessment: Unifying structure and texture similarity. IEEE Transactions on Pattern Analysis and Machine Intelligence  \textbf{44}(5),  2567--2581 (2022). \doi{10.1109/TPAMI.2020.3045810}

\bibitem{fan2024lightgaussian}
Fan, Z., Wang, K., Wen, K., Zhu, Z., Xu, D., Wang, Z.: Lightgaussian: Unbounded 3d gaussian compression with 15x reduction and 200+ fps. Advances in neural information processing systems  \textbf{37},  140138--140158 (2024)

\bibitem{fang2024mini}
Fang, G., Wang, B.: Mini-splatting: Representing scenes with a constrained number of gaussians. In: European conference on computer vision. pp. 165--181. Springer (2024)

\bibitem{franke2025vr}
Franke, L., Fink, L., Stamminger, M.: Vr-splatting: Foveated radiance field rendering via 3d gaussian splatting and neural points. Proceedings of the ACM on Computer Graphics and Interactive Techniques  \textbf{8}(1),  1--21 (2025)

\bibitem{girish2024eagles}
Girish, S., Gupta, K., Shrivastava, A.: Eagles: Efficient accelerated 3d gaussians with lightweight encodings. In: European Conference on Computer Vision. pp. 54--71. Springer (2024)

\bibitem{hanson2025speedy}
Hanson, A., Tu, A., Lin, G., Singla, V., Zwicker, M., Goldstein, T.: Speedy-splat: Fast 3d gaussian splatting with sparse pixels and sparse primitives. In: 2025 IEEE/CVF Conference on Computer Vision and Pattern Recognition (CVPR). pp. 21537--21546 (2025). \doi{10.1109/CVPR52734.2025.02006}

\bibitem{hanson2025pup}
Hanson, A., Tu, A., Singla, V., Jayawardhana, M., Zwicker, M., Goldstein, T.: Pup 3d-gs: Principled uncertainty pruning for 3d gaussian splatting. In: Proceedings of the Computer Vision and Pattern Recognition Conference. pp. 5949--5958 (2025)

\bibitem{10.1145/3272127.3275084}
Hedman, P., Philip, J., Price, T., Frahm, J.M., Drettakis, G., Brostow, G.: Deep blending for free-viewpoint image-based rendering. ACM Trans. Graph.  \textbf{37}(6) (Dec 2018)

\bibitem{kerbl3Dgaussians}
Kerbl, B., Kopanas, G., Leimk{\"u}hler, T., Drettakis, G.: 3d gaussian splatting for real-time radiance field rendering. ACM Transactions on Graphics  \textbf{42}(4) (July 2023), \url{https://repo-sam.inria.fr/fungraph/3d-gaussian-splatting/}

\bibitem{kim2024c3}
Kim, H., Bauer, M., Theis, L., Schwarz, J.R., Dupont, E.: C3: High-performance and low-complexity neural compression from a single image or video. In: 2024 IEEE/CVF Conference on Computer Vision and Pattern Recognition (CVPR). pp. 9347--9358 (2024). \doi{10.1109/CVPR52733.2024.00893}

\bibitem{10.1145/3072959.3073599}
Knapitsch, A., Park, J., Zhou, Q.Y., Koltun, V.: Tanks and temples: benchmarking large-scale scene reconstruction. ACM Trans. Graph.  \textbf{36}(4) (Jul 2017)

\bibitem{ladune2023coolchic}
Ladune, T., Philippe, P., Henry, F., Clare, G., Leguay, T.: Cool-chic: Coordinate-based low complexity hierarchical image codec. In: 2023 IEEE/CVF International Conference on Computer Vision (ICCV). pp. 13469--13476 (2023). \doi{10.1109/ICCV51070.2023.01243}

\bibitem{lee2025optimized}
Lee, J.C., Ko, J.H., Park, E.: Optimized minimal 3d gaussian splatting. In: The Thirty-ninth Annual Conference on Neural Information Processing Systems (2025), \url{https://openreview.net/forum?id=GMiC4ccyHn}

\bibitem{lee2024compact}
Lee, J.C., Rho, D., Sun, X., Ko, J.H., Park, E.: Compact 3d gaussian representation for radiance field. In: Proceedings of the IEEE/CVF Conference on Computer Vision and Pattern Recognition. pp. 21719--21728 (2024)

\bibitem{lin2025metasapiens}
Lin, W., Feng, Y., Zhu, Y.: Metasapiens: Real-time neural rendering with efficiency-aware pruning and accelerated foveated rendering. In: Proceedings of the 30th ACM International Conference on Architectural Support for Programming Languages and Operating Systems, Volume 1. pp. 669--682 (2025)

\bibitem{liu2024hemgs}
Liu, L., Chen, Z., Jiang, W., Wang, W., Xu, D.: Hemgs: A hybrid entropy model for 3d gaussian splatting data compression. arXiv preprint arXiv:2411.18473  (2024)

\bibitem{liu2024compgs}
Liu, X., Wu, X., Zhang, P., Wang, S., Li, Z., Kwong, S.: Compgs: Efficient 3d scene representation via compressed gaussian splatting. In: Proceedings of the 32nd ACM International Conference on Multimedia. pp. 2936--2944 (2024)

\bibitem{scaffoldgs}
Lu, T., Yu, M., Xu, L., Xiangli, Y., Wang, L., Lin, D., Dai, B.: Scaffold-gs: Structured 3d gaussians for view-adaptive rendering. In: 2024 IEEE/CVF Conference on Computer Vision and Pattern Recognition (CVPR). pp. 20654--20664 (2024). \doi{10.1109/CVPR52733.2024.01952}

\bibitem{mabyduck}
Mabyduck: Mabyduck. \url{https://www.mabyduck.com} (2025), last accessed 2026/03/04

\bibitem{mallick2024taming}
Mallick, S.S., Goel, R., Kerbl, B., Steinberger, M., Carrasco, F.V., De~La~Torre, F.: Taming 3dgs: High-quality radiance fields with limited resources. In: SIGGRAPH Asia 2024 Conference Papers. pp. 1--11 (2024)

\bibitem{filmgrainAV1}
Norkin, A., Birkbeck, N.: Film grain synthesis for av1 video codec. In: 2018 Data Compression Conference. pp. 3--12 (2018). \doi{10.1109/DCC.2018.00008}

\bibitem{papantonakis2024reducing}
Papantonakis, P., Kopanas, G., Kerbl, B., Lanvin, A., Drettakis, G.: Reducing the memory footprint of 3d gaussian splatting. Proceedings of the ACM on Computer Graphics and Interactive Techniques  \textbf{7}(1),  1--17 (2024)

\bibitem{qiu2024wasserstein}
Qiu, Y., Wagner, A.B., Ballé, J., Theis, L.: Wasserstein distortion: Unifying fidelity and realism. In: 2024 58th Annual Conference on Information Sciences and Systems (CISS). pp.~1--6 (2024). \doi{10.1109/CISS59072.2024.10480168}

\bibitem{ren2024octree}
Ren, K., Jiang, L., Lu, T., Yu, M., Xu, L., Ni, Z., Dai, B.: Octree-gs: Towards consistent real-time rendering with lod-structured 3d gaussians. IEEE Transactions on Pattern Analysis and Machine Intelligence pp. 1--15 (2025). \doi{10.1109/TPAMI.2025.3568201}

\bibitem{ren2025fastgs}
Ren, S., Wen, T., Fang, Y., Lu, B.: Fastgs: Training 3d gaussian splatting in 100 seconds. arXiv preprint arXiv:2511.04283  (2025)

\bibitem{VGG}
Simonyan, K., Zisserman, A.: Very deep convolutional networks for large-scale image recognition. In: Bengio, Y., LeCun, Y. (eds.) 3rd International Conference on Learning Representations, {ICLR} 2015, San Diego, CA, USA, May 7-9, 2015, Conference Track Proceedings (2015), \url{http://arxiv.org/abs/1409.1556}

\bibitem{NEURIPS2021_a11ce019}
Sitzmann, V., Rezchikov, S., Freeman, B., Tenenbaum, J., Durand, F.: Light field networks: Neural scene representations with single-evaluation rendering. In: Ranzato, M., Beygelzimer, A., Dauphin, Y., Liang, P., Vaughan, J.W. (eds.) Advances in Neural Information Processing Systems. vol.~34, pp. 19313--19325. Curran Associates, Inc. (2021), \url{https://proceedings.neurips.cc/paper_files/paper/2021/file/a11ce019e96a4c60832eadd755a17a58-Paper.pdf}

\bibitem{sun2024f}
Sun, X., Lee, J.C., Rho, D., Ko, J.H., Ali, U., Park, E.: F-3dgs: Factorized coordinates and representations for 3d gaussian splatting. In: Proceedings of the 32nd ACM International Conference on Multimedia. pp. 7957--7965 (2024)

\bibitem{Tseng_2026_WACV}
Tseng, Y.J., Kao, C.H., Chen, J.Z., Gnutti, A., Lo, S.Y., Lin, Y.Y., Peng, W.H.: Csgaussian: Progressive rate-distortion compression and segmentation for 3d gaussian splatting. In: Proceedings of the IEEE/CVF Winter Conference on Applications of Computer Vision (WACV). pp. 6883--6892 (March 2026)

\bibitem{wang2024end}
Wang, H., Zhu, H., He, T., Feng, R., Deng, J., Bian, J., Chen, Z.: End-to-end rate-distortion optimized 3d gaussian representation. In: European Conference on Computer Vision. pp. 76--92. Springer (2024)

\bibitem{wang2024contextgs}
Wang, Y., Li, Z., Guo, L., Yang, W., Kot, A., Wen, B.: Context{GS} : Compact 3d gaussian splatting with anchor level context model. In: The Thirty-eighth Annual Conference on Neural Information Processing Systems (2024), \url{https://openreview.net/forum?id=W2qGSMl2Uu}

\bibitem{SSIM}
Wang, Z., Bovik, A., Sheikh, H., Simoncelli, E.: Image quality assessment: from error visibility to structural similarity. IEEE Transactions on Image Processing  \textbf{13}(4),  600--612 (2004). \doi{10.1109/TIP.2003.819861}

\bibitem{zhan2025cat3dgs}
Zhan, Y.T., Ho, C.Y., Yang, H., Chen, Y.H., Chiang, J.C., Liu, Y.L., Peng, W.H.: {CAT-3DGS: A context-adaptive triplane approach to rate-distortion-optimized 3DGS compression}. In: Proceedings of the Thirteenth International Conference on Learning Representations (ICLR) (2025)

\bibitem{zhan2025cat}
Zhan, Y.T., Yang, H.b., Ho, C.Y., Chiang, J.C., Peng, W.H.: Cat-3dgs pro: A new benchmark for efficient 3dgs compression. In: 2025 33rd European Signal Processing Conference (EUSIPCO). pp. 1367--1371 (2025). \doi{10.23919/EUSIPCO63237.2025.11226320}

\bibitem{lpips}
Zhang, R., Isola, P., Efros, A.A., Shechtman, E., Wang, O.: The unreasonable effectiveness of deep features as a perceptual metric. In: 2018 IEEE/CVF Conference on Computer Vision and Pattern Recognition. pp. 586--595 (2018). \doi{10.1109/CVPR.2018.00068}

\bibitem{Zhang2025GaussianSpa}
Zhang, Y., Jia, W., Niu, W., Yin, M.: Gaussianspa: An "optimizing-sparsifying" simplification framework for compact and high-quality 3d gaussian splatting. In: 2025 IEEE/CVF Conference on Computer Vision and Pattern Recognition (CVPR). pp. 26673--26682 (2025). \doi{10.1109/CVPR52734.2025.02484}

\bibitem{zhou2025perceptualgs}
Zhou, H., Ni, Z.: Perceptual-{GS}: Scene-adaptive perceptual densification for gaussian splatting. In: Forty-second International Conference on Machine Learning (2025), \url{https://openreview.net/forum?id=ij0vj0BC72}

\end{thebibliography}

\end{document}